# AN ONTOLOGY-BASED APPROACH TO RELAX TRAFFIC REGULATION FOR AUTONOMOUS VEHICLE ASSISTANCE

Philippe Morignot, Fawzi Nashashibi
INRIA Rocquencourt, Team IMARA, Domaine de Voluceau, B.P. 105, 78153 Le Chesnay, France
Philippe.Morignot@inria.fr Fawzi.Nashashibi@inria.fr

**ABSTRACT**
Traffic regulation must be respected by all vehicles, either human- or computer- driven. However, extreme traffic situations might exhibit practical cases in which a vehicle should safely and reasonably relax traffic regulation, e.g., in order not to be indefinitely blocked and to keep circulating. In this paper, we propose a high-level representation of an automated vehicle, other vehicles and their environment, which can assist drivers in taking such "illegal" but practical relaxation decisions. This high-level representation (an ontology) includes topological knowledge and inference rules, in order to compute the next high-level motion an automated vehicle should take, as assistance to a driver. Results on practical cases are presented.

## 1. Introduction

Imagine that you are driving your car and that a truck is before you on the street, engine stopped, rear door open and unloading furniture for some close apartment. Since your car's lane is delimited by a continuous line and a sidewalk, you must not overtake according to the traffic regulation. You are then condemned to wait until the truck has finished unloading, a process which might keep you stopped for an uncertain, probably long, amount of time.

To take a second example, imagine that you are about to reach a roundabout, but that the car before yours on the lane has stopped, probably with an engine problem, e.g., electric power cut. Here again, since this lane is delimited by a continuous line and a sidewalk, strict respect of traffic regulation condemns you to wait behind the defective car until that car can move again, a process which might probably be counted in hours.

Many similar practical situations can be imagined, or taken from every driver's experience.

Human drivers can cope with such abnormal situations. For example, after having waited for some amount of time, a human driver might decide to cross the continuous line: He checks for the absence of vehicles on the adjacent opposite lane, makes a small left turn, crosses the continuous line, overtakes the unloading truck or the defective car, drives a few meters on the adjacent lane, and comes back to its initial lane once the obstacle is passed. Alternatively, the driver could decide to slowly run on the sidewalk to overtake the stopped truck / defective car.

Strictly speaking, traffic regulation is violated indeed: The French road traffic regulation states that "*vehicles must circulate on roadways, except in case of absolute emergency*" (section R412-7 [8] for France, [6] for an international definition). But in practice, given the above unusual circumstances, no one will blame a driver for safely crossing the above continuous line after having waited for a reasonable amount of time. Perhaps even a policeman, if present, would evaluate the abnormality of the situation and would impose you to cross this continuous line and overtake the stopped truck/vehicle. In other words, perfect traffic respecting perfect regulation in a perfect world is not the way things happen in real open environments.

If the above decision can be taken every day by human drivers, the picture is different for an autonomous vehicle driven by a computer. In the two above situations, an intelligent robotic vehicle such as a CyCab [7], based on perception and control, will stop and be kept stuck on its lane until the unloading truck/damaged vehicle moves. Whereas, to mimic human behavior, a decision would be needed at some point: should the CyCab follow its obstacle avoidance algorithm, i.e., change lane, or should it follow traffic regulation, i.e., stay on its lane? (In the experiments on the CityMobil project in the city LaRochelle in Nov. 2011, kids were having fun with a CyBus by stepping in front of it each time it avoided them and started to run.)

The purpose of this paper is to give to autonomous vehicles such as a CyCab enough reasoning capabilities to be able to take such decision, and therefore be able to cope with such unusual situations. In other words, not letting the autonomous vehicle be stuck in unusual but practical situations, such as the two above, because of (probably overly restrictive) traffic regulation following. More generally, such reasoning capability is one aspect of decisional autonomy for vehicles, which is considered as a major research area of this century towards intelligent traffic [13].

The paper is organized as follows: First, a model based on an ontology including rules is presented in section 2. Second, an implementation based on the ontology editor PROTÉGÉ [17] and SWRL (Semantic Web Rule Language) [11] is presented, and results are described, in section 3. Finally we relate our work to previous approaches and sum up our contribution.

## 2. Model

### 2.1 Context

Among the possible approaches to modeling traffic situations, there is an increasing number of symbolic ones using ontologies [3] [12] [18]. The main idea is that a high level, symbolic, representation (knowledge) is useful to perform reasoning on traffic situations, as complementary to low level ones involving perception, or path planning and kinematic control --- see [1] on collaborating vehicles integrating these two approaches. More generally, ontologies are introduced into mobile robotic frameworks (e.g., OROCOS [16]).

Other approaches use vehicular ad hoc networks (VANETs) in order to model sensor and actuation inside each vehicle, and communication among them (V2V) or with the infrastructure (V2I) [14] [15]. In our approach, we also use vehicles equipped with sensors and actuators, and which can communicate with others and the infrastructure. But we focus on the internal part of each vehicle only, and specifically its decisional part, instead of building statistics over the global traffic as a whole (how dense it is and how to reduce it).

### 2.2 Ontologies

In Computer Science (as opposed to Philosophy, where the term "ontology" has a different definition), an ontology may be defined as a specification of a conceptualization of a domain of knowledge [10]. For example, infectious disease diagnosis is a domain of medical knowledge; The concepts involved, lying in the brain of physicians, constitute its conceptualization; And a description of these concepts in a formal language constitutes its specification.

An important characteristic of an ontology is its completeness: an ontology should completely cover a knowledge domain, i.e. not leaving concepts behind. An ontology may also be defined as a complete semantic network, emphasizing that it is composed of a hierarchy of concepts.

In practice, an ontology is expressed as classes, properties and individuals. Tools are available to graphically create/edit an ontology (e.g., PROTÉGÉ [17], SWOOP [20]) and express it in OWL (Ontology Web Language).

### 2.3 Proposed model

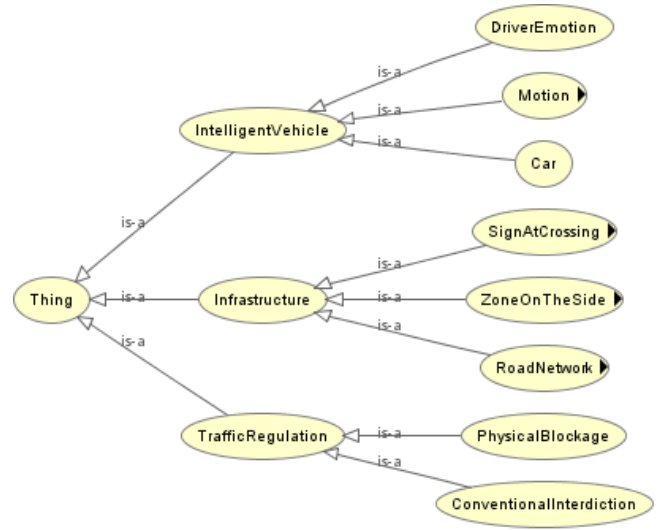

Fig. 1: Toplevel of the class hierarchy in the ontology, as shown by Graphviz.

The proposed ontology represents the vehicle (the intelligent vehicle and other vehicles), the infrastructure and the traffic regulation (see Fig. 1). Aiming at modeling all the concepts involved in traffic regulation relaxation, we found no existing ontology dedicated to it (e.g., A3ME focuses on the vehicle's motion only [3]) --- only newspapers articles report accidents in case of traffic regulation relaxation. Therefore, we built our own ontology, not based on a texts corpus, but on drivers' experience (member of the lab with their driving license) and their own reactions regarding traffic regulation relaxation.

More precisely, a vehicle is symbolically represented as its name ("*Car*" class), an internal class ("*DriverEmotion*") and its possible motion ("*Motion*" class, with sub-classes "*CurrentMotion*" and "*NextMotion*"). The main object properties to represent an intelligent vehicle are its motion and its location.

The above static representation is completed by a dynamic one using inference rules (see section 3.2). These rules are used to augment the static description, i.e., make inferences based on the knowledge of the situation at hand. In practice, an inference rule adds object properties or determine a class of an individual.

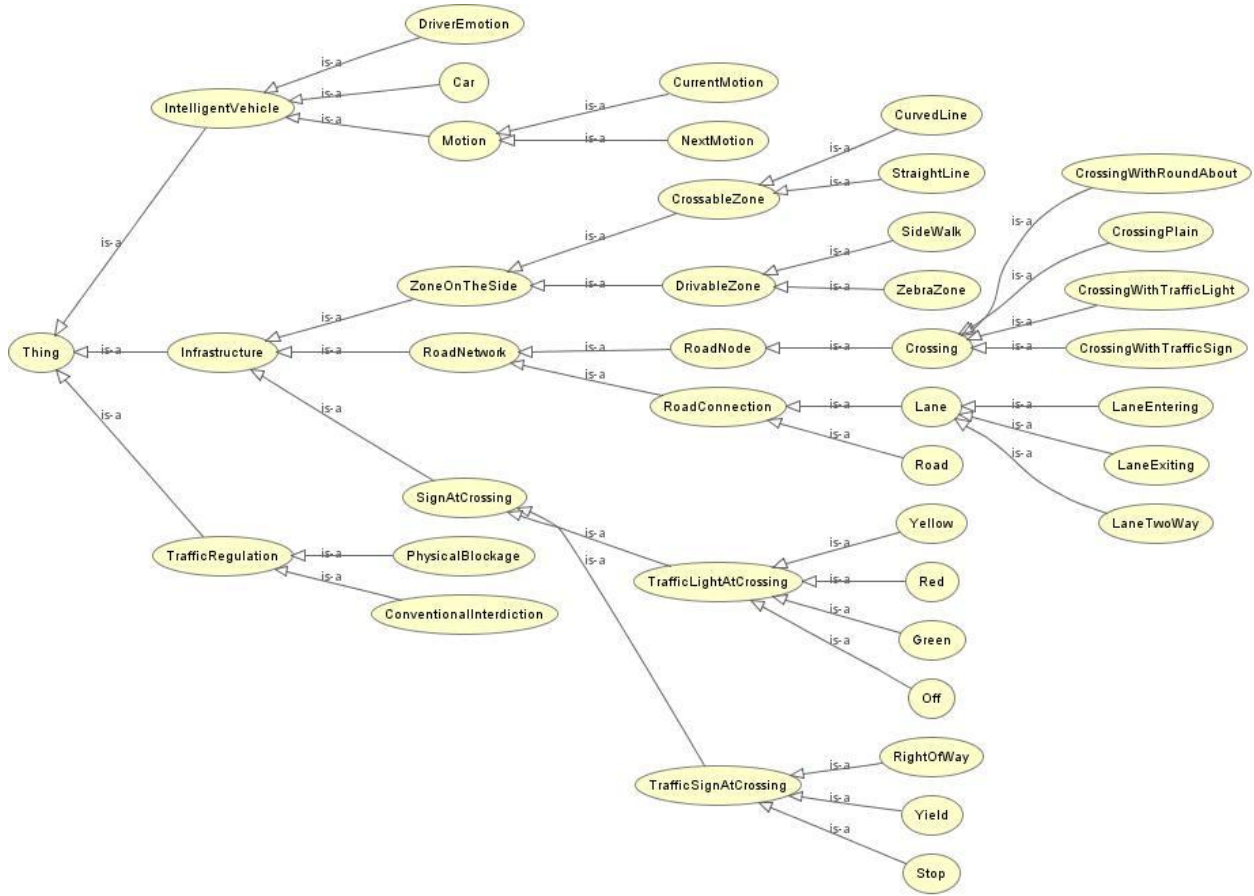

Fig. 2: Ontology for relaxing traffic regulations, as represented by GraphViz.

In our representation, two main properties of a vehicle must be inferred: the "*isOn*" property specifies an individual of the infrastructure on which a vehicle is (e.g., a named lane); and "*hasMotion*" / "*hasNextMotion*" object properties relate a vehicle to an individual of the "*Motion*" class. For now, the possible motions of a vehicle are set by individuals of this class (see Fig. 1):"*Forward*", "*Stopped*" and "*Backward*" --- this discretization of the speed of a vehicle can be made as precise as desired, e.g., with integers representing a value in km/h.

The traffic regulation (the bottom class in Fig. 1) is represented by individual related by an object property "*isIllegal*" to the motion of the intelligent vehicle.

Being a high-level topological model, our approach relies on symbols to describe an intelligent vehicle, other vehicles and their environment. Therefore, as in [3] [12] [18], we have to assume that (1) perception is capable of turning sensor data into symbols; and, conversely, that (2) the inferred symbolic motion leads to path planning and kinematic control, for actual motion --- these two areas of robotics research (perception and control) are out of the scope of this paper.

## 3. Results

The approach in Section II is implemented with the ontology editor PROTÉGÉ version 4 [17], using the reasoner PELLET version 2.2.0. Inference rules are expressed in SWRL (Semantic Web Rule Language) [11]. We first present the static knowledge involved (i.e., classes, properties and individuals), then present the dynamic knowledge (i.e., inference rules) and finally present examples.

### 3.1 Classes, properties and individuals

The environment of the CyberCar is represented by the class "*Infrastructure*" (see Fig. 2). The main representational element is the "*RoadNetwork*" class, representing lanes/roads and intersections as a graph, with the former being vertices and the latter being edges (borrowed from [18]). For example, two intersections (i.e., "*RoadNode*" individuals) can be connected to another intersection by one road (i.e., one "*RoadConnection*" individual), or even by two lanes (i.e., two "*RoadConnection*" individuals), the lane up and the lane down --- a two-way lane connects the two intersections.

The immediate delimiters of each "*RoadConnection*" individual are represented by the "*ZoneOnTheSide*" class, e.g., sidewalks, zebra zones, continuous lines and dashed lines. The first two are sub-class of the "*DrivableZone*" class, i.e., at the detail level of description, they occupy some surface on the ground, hence can be physically driven on by a vehicle. The last two also are sub-class of the "*CrossableZone*" class, i.e., at the considered level of description they do not occupy any surface on the ground, hence can physically be crossed only, not driven on. To follow the example of the previous paragraph, the two lanes (up and down) of a two-way road (a "*RoadConnection*" individual) can be separated by a continuous line, i.e., an individual of the "*ContinuousLine*" class, sub-class of the "*CrossableZone*" class. The "*hasBesides*" object property links "*RoadConnection*" individuals to "*ZoneOnTheSide*" ones.

The last class in Fig. 2, "*SignAtCrossing*", represents traffic lights and signs at an intersection, i.e., a "*RoadNode*" individual. Individuals of that class can be used to infer conflicts among vehicles approaching an intersection, hence to infer the right-of-way of a vehicle arriving on a "*RoadConnection*" individual connected to a "*RoadNode*" individual (see [18] for a discussion on this point).

**3.2 Inference rules**
The evolution of a symbolic situation should be represented with terms which are sometimes true and sometimes false, depending on the time at which they are observed (a flavor of *fluents* in STRIPS task planning [9]). But, although the "not" operator, negating a term, can be represented in OWL to some extent (see [17]), Description Logic (DL), the formal basis of ontologies, is monotonic and is not capable of representing the new true/false value of a term, the truth value of which changes. Typically, SWRL [11] cannot represent a rule: **IF** A **AND** B **AND** C **THEN** not(D). That is, in DL, things can only be added to the current ontology, and never retracted from it.

Facing this restriction, we chose to discretize time, i.e. reason on time steps, and represent the reasoning of an intelligent vehicle over two successive time steps only: inferring the next motion of a vehicle, the current motion and context being symbolically described. Then, the inference mechanism over our ontology, computing this next motion, is supposed to be iterated over time (with an update phase interleaved), in order to build the long term course of action of an autonomous vehicle. Formally, the ontology is used as a mapping: $S \times M \rightarrow M$ where $S$ is the set of situations, one situation being expressed by an ontology, and $M$ the set of motions of an autonomous vehicle. The situation $s_n \in S$ and motion $m_n \in M$ at time step $n$ produce the motion $m_{n+1} \in M$ at time step $n+1$.

CrossableZone(?s), Car(?a), Car(?b),

Lane(?l1), Lane(?l2),

hasEmotion(?a, Nervous),

isAfter(?a, ?b),

hasBesides(?l1, ?s), hasBesides(?l2, ?s),

hasMotion(?a, Stopped),

isOn(?a, ?l1), isOn(?b, ?l1), DifferentFrom (?l1, ?l2),

isIllegal(?l1, ?l2),

isClear(?l2)

->

isNextOn(?a, ?l2)

> Table 1: An example of SWRL inference rule in the case of two lanes separated by a thin delimiter.

DrivableZone(?s), Car(?a), Car(?b), Lane(?l),

hasEmotion(?a, Nervous),

isAfter(?a, ?b),

hasBesides(?l, ?s),

hasMotion(?a, Stopped),

isOn(?a, ?l), isOn(?b, ?l),

isIllegal( ?l, ?s),

->

isNextOn(?a, ?s)

Table 2: An example of SWRL inference rule in the case of one lane with a unique large delimiter.

Inference rules express how to relate the current situation and motion of a vehicle to its next motion. We use such inference rules to actually encode the traffic regulation relaxation behavior of one specific vehicle. These rules strongly participate in attaining the resulting behavior we initially targeted for the intelligent vehicles, so we describe some of these rules now.

The rule in Table 1 expresses that if vehicle *?a* on lane *?l1* is behind the stopped vehicle *?b* on the same lane, then *?a* can cross the lane delimiter *?s*, even if it is illegal, to reach the adjacent opposite lane *?l2*, provided that it is clear (no vehicle on it). The representation choices in this rule lead to several comments:

1. The term *isClear(?l)* expresses that there is no vehicle on lane *?l*. In first order predicate logic, it would be written as: $\forall c \in Car, \forall l \in RoadConnection$ :
   $on(c, l) \Rightarrow \neg ( \exists c' \in Car, c \neq c' \land on(c', l) )$

   (A variant includes an additional term: $distance(c,c') < |\ speed(c) - speed(c')\ | * T_{overtake}$). Unfortunately, such negated existential quantification in the second term of the above implication cannot be expressed in DL. As a first approach, we chose to encode the clearness of a lane as a class "isClear", its individuals being the current clear lanes --- another mechanism is needed in order to maintain these individuals in this class.

2. The property *isIllegal(?l1, ?l2)* expresses that moving from RoadConnection individual *?l1* to RoadConnection or ZoneOnTheSide individual *?l2* is not legal given traffic regulation. As such, it should be inferred from a representation of the traffic regulation [12] [18], e.g., rules concluding on the legality/illegality of a given motion. As a first approach, we chose a simpler implementation: enumerating a set of traffic regulation violations with the *isIllegal* property.
3. The waiting time of the vehicle *?a* behind the front stopped vehicle *?b* is represented by the property *hasEmotion(Car, DriverEmotion)*. If the so-called "driver emotion" of car *?a* is "Nervous", then the waiting time is considered to have expired and the illegal motion can be performed. If another "driver emotion", e.g., "Relaxed", is active, then the waiting time is considered to still run, therefore the vehicle *?a* keeps being stopped behind front vehicle *?b* (the corresponding inference rule is shown in Table 3). Another mechanism is needed to connect the time elapsed since vehicle *?a* is blocked behind vehicle *?b*, to the individuals "Nervous" or "Relaxed".

The rule in Table 2, complementary to the one of Table 1, encodes the case of a one-way lane with a unique large delimiter (e.g., a sidewalk).

Other rules do not compute the next motion of an intelligent vehicle, but fill in the gaps in the current situation of the intelligent vehicle. For example, the rule of Table 4 concludes on the "*hasMotion*" object property of a car (motion at time step *n*), and not on the "*hasNextMotion*" one (motion at time step *n+1*).

The rule in Table 5 is a default one, expressing that intelligent vehicle *?b* runs normally when there is no obstacle stopping it (regular case).

> Car(?a), Car(?b),
>
> hasMotion(?a, Stopped),
>
> isBefore(?a, ?b)
>
> ->
>
> hasMotion(?b, Stopped)

Table 4: An example of SWRL inference rule, to prevent a car from colliding another car stopped in front of it.

> Car(?a), Car(?b),
>
> hasNextMotion(?a, Forward),
>
> isBefore(?a, ?b)
>
> ->
>
> hasNextMotion(?b, Forward)

Table 5: An example of SWRL inference rule, for the regular case.

### 3.3 Examples

The first case of section 1 is depicted in Fig. 3.

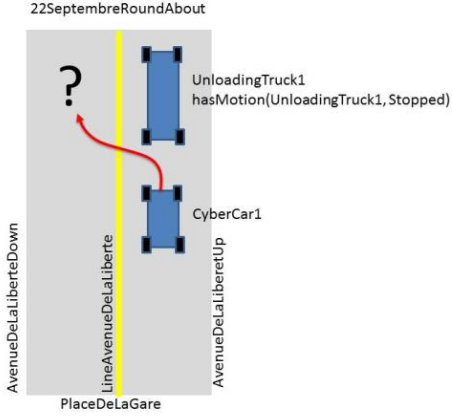

Fig. 3: First case of section I, a truck is unloading in front of the intelligent vehicle.

In this case, the infrastructure is composed of a roundabout "*22SeptembreRoundAbout*" connected by a two-way lane ("*AvenueDeLaLiberteUp*" and "*AvenueDeLaLiberteDown*" individuals) to a place ("*PlaceDeLaGare*"). A ZoneOnTheSide individual ("*LineAvenueDeLaGare*" individual) is between the two ways of this lane (represented by the "*hasBesides*" object property). The stopped unloading truck is represented by "*Car*" individual "*UnloadingTruck1*" and is stopped (object property "*HasMotion(UnloadingTruck1, Stopped)*"). A second vehicle is the CyCab (individual "*CyberCar1*" of the class "*Car*"). Both vehicle stays on the "*AvenueDeLaLibertUp*" lane through the object property "*isOn*".

Here are the inferred object properties:

hasMotion(CyberCar1, Stopped)                                    (1)

isAfter(CyberCar1,UnloadingTruck1)                               (2)

hasNextMotion(CyberCar1, Forward)                                (3)

isNextOn(CyberCar1, AvenueDeLaLibertDown)                        (4)

The first two inferred object properties fill in the current situation: the inferred property 1 is a result of the firing of the rule in Table 4; the inferred property 2 is performed because the "*isBefore*" and "*isAfter*" properties are declared inverse. The last two inferred object properties describe the next motion of the individual CyberCar1: the object property 3 is the result of firing a rule close to the ones of Table 1 and 2, but related to the speed of the vehicle; the object property 4 is the result of the firing of the rule in Table 1.

Interestingly, the above resulting behavior of the intelligent vehicle could not be reached if the so called "DriverEmotion" was "Relax". Here is the inferred object properties in that case:

hasMotion(CyberCar1, Stopped)                                    (1)

isAfter(CyberCar1,UnloadingTruck1)                               (2)

hasNextMotion(CyberCar1, Stopped)                                (3)

That is, under a different DriverEmotion, the vehicle CyberCar1 is stopped behind vehicle UnloadingTruck1 (as in the previous example), but now his next motion will keep being "Stopped", i.e., the CyberCar waits behind the unloading truck, as respect of traffic regulation prescribes. This is the default behavior of intelligent vehicles, and our whole approach results in crossing the continuous line (a behavior prohibited by the traffic regulation) in this example of unusual circumstances.

The second example of section I is close in spirit to the example of Fig. 3 (it only differs from it by the topology of RoadConnection and RoadNode individuals). Here are the inferred object properties:

hasMotion(CyberCar2, Stopped)                                    (1)

isAfter(CyberCar2,UnloadingTruck2)                               (2)

hasNextMotion(CyberCar2, Forward)                                (3)

isNextOn(CyberCar2, SwRueDu22Septembre)                          (4)

The main difference with the first example is that "*CyberCar2*" individual is next on a sidewalk to overtake the unloading truck ("*SwRueDu22Septembre*" individual), since the inference rule of Table 2 has fired.

The reasoner PELLET performs the above inferences on these two cases together in 389 ms, on a machine 4-core at 2 GHz with 4 Gb RAM. But some time in this figure is spent classifying the ontology, i.e., sorting the classes along the "*is-a*" relation and checking them for consistency (i.e., every class is not prevented from owning individuals). Other reasoners are available (e.g., FACT++, RACERPRO), and using benchmarks' results for choosing on another reasoner can improve these performances [4].

## 4. Discussion

Defining a topological world model to infer the next motion of an intelligent vehicle to assist drivers regarding traffic regulation relaxation raises numerous issues:

1. How long does a driver take, facing the situations described in section I, to decide to cross a continuous line? On one side, crossing this line is forbidden by traffic regulation (the goal being to respect traffic regulation), but on the other side staying too long trapped in his lane behind an unloading truck/defective vehicle seems inappropriate either (the goal being to keep circulating, e.g., reaching point B from point A). The time it takes to overtake an unloading truck/defective vehicle is related to the way a driver finds an acceptable trade off to this conflict --- this could be encoded as a driver-dependent threshold on the elapsed time, regarding the "*hasEmotion*" property of section 3.2. But this question is relevant to the domain of cognitive psychology (e.g., see [2]), which is out of the scope of this paper.

2. A drawback of an ontology-based approach is that a vehicle and its environment are represented in discrete, symbolic terms: things are true or false but there is no way to represent something intermediate, i.e., a notion of uncertainty (uncertainty is implicit in OWL because of the open world assumption, stating that if a term is not present in the ontology, it is assumed to be unknown, as opposed to the closed world assumption in task planning [9]). For example, Bayesian networks can represent probabilistic inferences, i.e., reason on the uncertainty inherent to the involved terms (which would be called *state variables*). Therefore a first solution to representing uncertainty would be to re-write the above rules (see section 3.2) as probabilistic dependencies among state variables. A second solution is to restrict our view to describing the intelligent vehicle's context only, i.e., providing the right ontology for the current context, and making inferences with certainty about it. Further reasoning, including uncertainty inside these certainty limits, being performed by Bayesian networks.

3. Regele [18] and Hulsen et al. [12] use a high level topological representation of the environment to make inferences about the conflicts at an intersection (e.g., giving right-of-way). But if we use a high level topological representation too, these bodies of work stay close to the traffic regulation. That is, they infer with certainty properties of vehicles' possible motions given what is permitted by traffic regulation (a vehicle passes or does not pass). Our approach differs from theirs, in that it is closer to the vehicle's motion (e.g., see the "*DriverEmotion*" class in section 3) with a more detailed representation of vehicles, and dedicated to relaxing traffic regulation for practical purpose.

4. Mohandas et al. [14] propose a proportional integral controller to manage the congestion of traffic in vehicular ad hoc networks (VANETs). But there is no model of the vehicle, except in the queue at each VANET node. A close view is due to Mohimani et al. [15] in which a vehicle is a state automaton, and a probabilistic model is used to represent traffic in vehicular ad hoc networks. The closest part to our work is the state automaton representing the decisional part of a vehicle (e.g., answering the question: overtaking or not?). But we focus much more deeply on the decisional part with an ontology, to represent necessary knowledge to break or keep traffic regulation.

5. Other authors focus on emergency vehicles having de facto priority over regular traffic [3] [19]. But this is dedicated to specific vehicles, with little decision taken from it --- as opposed to them, we elaborate on the decisional part of each regular vehicle facing unusual traffic situations. Interestingly, Bermejo et al. [3] also use an ontology to represent the motion of regular vehicles (e.g., having to change lane to give free of way to an emergency vehicle). If this approach is probably the closest to ours, we represent in the ontology the whole infrastructure in which regular vehicles are embedded, and not only the motion parameters of each vehicle --- which can obviously be refined as deeply as desired in our model.

6. As stated earlier, the proposed approach relies on symbols (an ontology) to draw a decisional component into a vehicle equipped with sensors and effectors and potentially communicating with other vehicles and with an infrastructure, such as CyberCars [7]. As such, we envision to include it into the perception / planning / control cycle, after perception (which extracts symbolic information from signals of sensors) and before the path planning part (which computes a trajectory to reach a desired location in the current environment, sending low level commands to actuators). That is, symbolic information are available for that component and that component produces new goal locations, which could not be planned without the traffic relaxation module --- the remaining modules plan for respecting traffic regulation.

## 5. Conclusion

In this paper, we have presented a topological world model, to relax traffic regulation in unusual but practical situations, in order to assist drivers. This model is composed of an ontology, representing the vehicles, the infrastructure and the traffic regulation (implemented in OWL with Protégé [17]), and inference rules (implemented in SWRL [11]) computing the next motion of an intelligent vehicle under discretized time. Traffic regulation relaxation cases have been presented, exhibiting realistic behavior from the intelligent vehicle.

Future work involves (1) representing traffic regulation as rules inferring on the legality/illegality of an intelligent vehicle's potential motion; (2) integrating the ontology, a reasoning paradigm on certainty, as context definition for perception/control using uncertainty; and (3) porting the ontology on CyCab platforms.

## Acknowledgement(s)

The authors thank our colleagues of the IMARA team for numerous fruitful discussions. We thank Timothy Redmond (Stanford) for help on Protégé, and anonymous reviewers for helpful comments.